# Multi-scale Semantic Prior Features Guided Deep Neural Network for Urban Street-view Image Inpainting

Jianshun Zeng, Wang Li, Yanjie Lv, Shuai Gao, YuChu Qin

***Abstract*—Street-view image has been widely applied as a crucial mobile mapping data source. The inpainting of street-view images is a critical step for street-view image processing, not only for the privacy protection, but also for the urban environment mapping applications. This paper presents a novel Deep Neural Network (DNN), multi-scale semantic prior Feature guided image inpainting Network (MFN) for inpainting street-view images, which generate static street-view images without moving objects (e.g., pedestrians, vehicles). To enhance global context understanding, a semantic prior prompter is introduced to learn rich semantic priors from large pre-trained model. We design the prompter by stacking multiple Semantic Pyramid Aggregation (SPA) modules, capturing a broad range of visual feature patterns. A semantic-enhanced image generator with a decoder is proposed that incorporates a novel cascaded Learnable Prior Transferring (LPT) module at each scale level. For each decoder block, an attention transfer mechanism is applied to capture long-term dependencies, and the semantic prior features are fused with the image features to restore plausible structure in an adaptive manner. Additionally, a background-aware data processing scheme is adopted to prevent the generation of hallucinated objects within holes. Experiments on Apolloscapes and Cityscapes datasets demonstrate better performance than state-of-the-art methods, with MAE, and LPIPS showing improvements of about 9.5% and 41.07% respectively. Visual comparison survey among multi-group person is also conducted to provide performance evaluation, and the results suggest that the proposed MFN offers a promising solution for privacy protection and generate more reliable scene for urban applications with street-view images.**

## I. INTRODUCTION

Street-view images contain rich details of urban scenes and are important carriers for the effective expression of geographic information. In recent years, optical sensors and image understanding approaches powered by Artificial Intelligence (AI) have been widely adopted for data acquisition and processing in urban environments, especially with the rapid development of autonomous driving (AD) vehicles, video surveillance, online mapping services, the data volume of street-view images has experienced exponential growth. The large volume datasets facilitated the development of learning-based AI algorithms, including perception, scene parsing[4, 5], and planning[6, 7]. Although the rich datasets are useful in various applications such as environment management and urban planning, there are still major issues, e.g. moving object occlusion and privacy concerns in public areas, which would significantly reduce the data quality and availability in real applications. The privacy issue means that it is necessary to protect sensitive information (e.g., human faces, vehicle license plates) in public areas, the occlusion issue is related to the availability of image data in mapping services, as the region of interest may be occluded by distracting objects. The traditional approach to pre-processing involves applying blur to sensitive patches within street-view images. However, this method neglects the challenge posed by occlusion from moving objects. An integrated approach is greatly desired, as it can effectively eliminate both identity-related information and moving objects from street-view images in tandem. This results in images with minimized privacy implications and intermittent objects, which can then be utilized for further applications.

Image inpainting aims to effectively restore missing regions of damaged images with relevant visual content, which has been widely applied in various vision tasks, including image editing, object removal, and old photo restoration[8]. The image inpainting algorithms either fill the hole by matching similar patches within the image using rule-based methods[9] or synthesize missing content by training on a large corpus of images in learning-based methods[10-13]. Recently, the rising of deep feature learning[14] and generative adversarial learning[11] has significantly advancing the field of image inpainting. This process has led to the creation of visually plausible and impressive results on various datasets (e.g., natural images and human faces). Image inpainting approaches would be promising solutions to solve the issues by replacing the areas of moving objects with a realistic background.

Generally, street-view datasets capture images in diverse and intricate urban scenes with different spatial structures, objects, and environmental settings. The complexities of street-view

Manuscript submitted September 03, 2023; This work was supported by Hundred Talents Program of the Chinese Academy of Sciences, E03302030D.
*(Corresponding author: Yuchu Qin.)*

Jianshun Zeng is with Key Laboratory of Digital Earth Science, Aerospace Information Research Institute, Chinese Academy of Sciences, Beijing 100094, China, also with the School of Electronic, Electrical and Communication Engineering, University of Chinese Academy of Sciences, Beijing 100049, China (e-mail: zengjianshun21@mails.ucas.ac.cn).

Wang Li is with the State Key Laboratory of Remote Sensing Science, Aerospace Information Research Institute, Chinese Academy of Sciences, Beijing 100094, China (e-mail: lwwhdz@sina.com).

Yanjie Lv, Shuai Gao and Yuchu Qin are with International Research Center of Big Data for Sustainable Development Goals, Beijing 100094, China(e-mail: lvyj@aircas.ac.cn ; gaoshuai@aircas.ac.cn; qinyc@radi.ac.cn).

introduce difficulties in understanding urban scenes for object detection, semantic information extraction, and incorporating the features to restore occluded regions effectively as well. Fundamentally, understanding the global context through the undamaged regions and generating fine local details that align with the global scenes are the most crucial prerequisites for image inpainting. Existing networks adopted strategies, to perform image inpainting, e.g., encoder-decoder architectures[10, 15], dilated convolutions[16], gated convolutions[17], Fourier convolutions[3], and stylized transformer blocks[18], which simultaneously expand the receptive field and capture long-range dependencies. Furthermore, multi-stage approaches[19] utilize structural information, such as edges or labor-intensive semantic maps, to enhance image content reasoning. However, these approaches cannot effectively capture multi-level semantics in the unmasked region and propagate the image semanteme to the masked areas in complex urban scenes. As shown in Fig. 1(c-e), the incoherent local details would be generated with the existing models, potentially resulting in the presence of visual artifacts in street-view images.

Simultaneously, generating pairs of image, comprising an input image and a corresponding binary mask, is also a key step in model training for image inpainting algorithms. The objective of image inpainting models is to remove the moving objects by filling the areas with textures and structures of the background, while the new moving objects should not be hallucinated inside the images. Many existing inpainting approaches[20] rely on randomly selecting images without filtering scenes containing traffic crowds and generating masks by randomly locating rectangles or providing the area of moving objects directly. However, training models with such sample images would induce undesirable color bleeding along object boundaries, and also introduce the synthesis of object-like artifacts within the inpainted regions[21].

In this paper, we present a Multi-Scale Prior Feature Guided Image Inpainting Network (MFN) model, a novel Deep Neural Network that effectively restore the coherent structure and intricate details of static urban scenes, following the removal of moving objects as shown in Fig. 1(f). The network is designed with a semantic prior prompter with Semantic Pyramid Aggregation (SPA) modules and a semantic enhanced image generator with Learnable Prior Transferring (LPT) modules, contextual attention mechanism is also applied in the model. A background-aware data processing scheme was adopted to prevents the generator from synthesizing new objects inside holes and meets the removal application in street-view images.

In summary, our contributions are summarized as follows:

1.We propose MFN, a novel dual Encoder-Decoder architecture for street-view image inpainting, which formed by a semantic prior prompter with SPA blocks and a semantic enhanced image generator with LPT blocks and a contextual attention mechanism.

2. A background-aware data processing pipeline prevents the generator from synthesizing new objects inside holes and meets the removal application in street-view images.

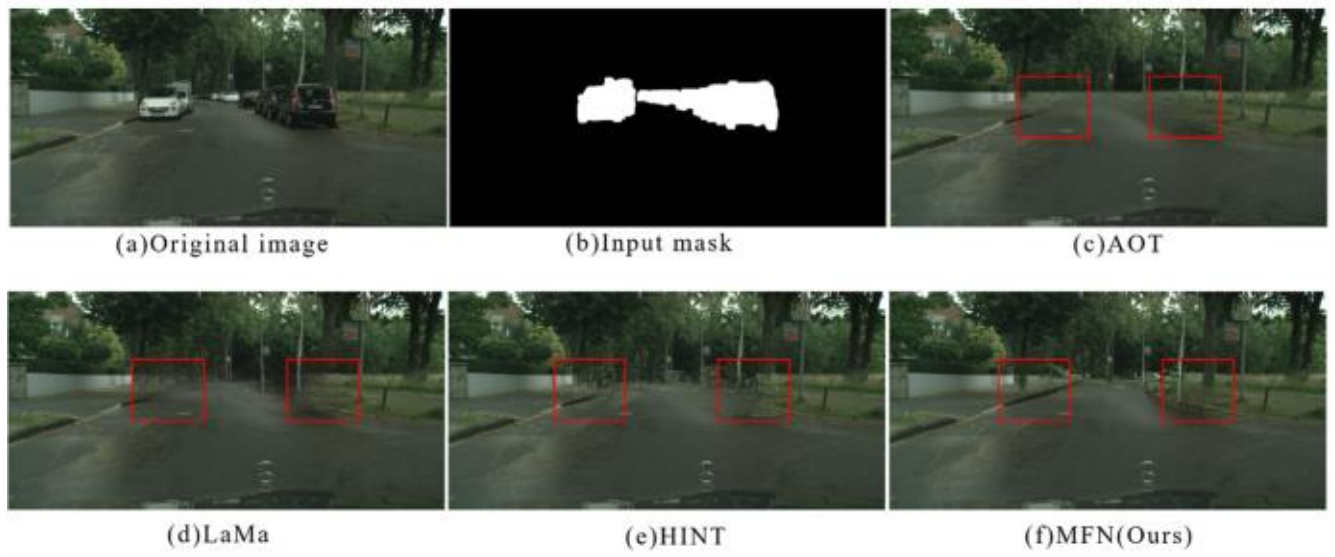

Fig. 1. Static urban scene image restoration of different methods. (a) and (b) are input image and mask. (c)Conventional techniques for street-view image (e.g., EC[1]) often rely on structural information such as object contours in multi-stage modules that generate unreliable results and boundaries. (d) and (e) are the inpainted results of the current high-resolution image inpainting method(e.g., AOT [2], LaMa [3]), which cannot retain original structural details. (f) Our method adopts rich knowledge from the pretrain model to obtain reasonable structure details to generate excellent results over the whole image.

3. Extensive experiments on the benchmark dataset demonstrate the superiority of our MFN in recovering visually realistic street-view images.

The remaining of this paper is organized as follows: Section 2 presents related works to the proposed approach. The proposed network is presented and described in Section 3. Section 4 gives the whole experimental details, results and discussion, including qualitative, quantitative and visual quality comparisons, and ablation experiments are introduced in this part. Finally, Section 5 contains the conclusions.

## II.RELATED WORK

Recently, several image inpainting strategies have been proposed to improve visual characteristics in images. These methods can be classified into two categories: rule-based methods and data-driven methods.

### *A. Rule-based Methods*

Generally, rule-based image inpainting algorithms could be broadly categorized into two groups: diffusion-based[9, 22] and patch-based[23, 24]. Diffusion-based methods aim to iteratively propagate visual content from neighboring known areas to the damaged regions, guided by specific partial differential equations. Patch-based methods assume that missing regions can be filled with appropriate candidate patches from undamaged regions or external databases. While traditional image inpainting methods have shown promising results for images with plain backgrounds or repeated textures, they face challenges in generating realistic content for complex scenes. This challenge primarily arises from their constrained semantic comprehension when confronted with areas of absence.

### *B. Data-Driven Methods*

In recent years, significant progress has been made by the emergence of Convolutional Neural Networks (CNNs)[25] and Generative Adversarial Networks (GANs)[26] for image inpainting. In contrast to traditional image inpainting methods, learning-based approaches can learn high-level features and image distributions that align with human visual perception through data-driven approaches. These approaches have been shown to improve the poor performance of traditional methods on real-world data in various scenarios. Pathak et al.[10] introduced an encoder-decoder architecture to perform global image context encoding and feature decoding. Subsequently, various U-Net based[27] and other notable networks have been proposed for image inpainting, aiming to reduce visual artifacts and generate high-quality images. Iizuka et al.[28] proposed global and local discriminators based on fully convolutional networks (FCN) to enhance the consistency of global and local content. Yu et al.[29] incorporated a context attention mechanism to find similarities for enhancing the restoration of high-frequency information. Following that, techniques such as partial convolution [13], gated convolution[17], and region normalization[29] were developed to improve the distinction between valid pixels and damaged regions during feature extraction. Furthermore, Suvorov et al.[3] introduced the Fast Fourier Convolution into image inpainting to achieve a larger effective receptive field for understanding the global structure of an image. Li et al.[18] developed a stylized transformer block for generating diverse image contexts. Despite the adoption of diverse network architectures and learning techniques, these methods encounter challenges in maintaining satisfactory structural consistency in complex scenes.

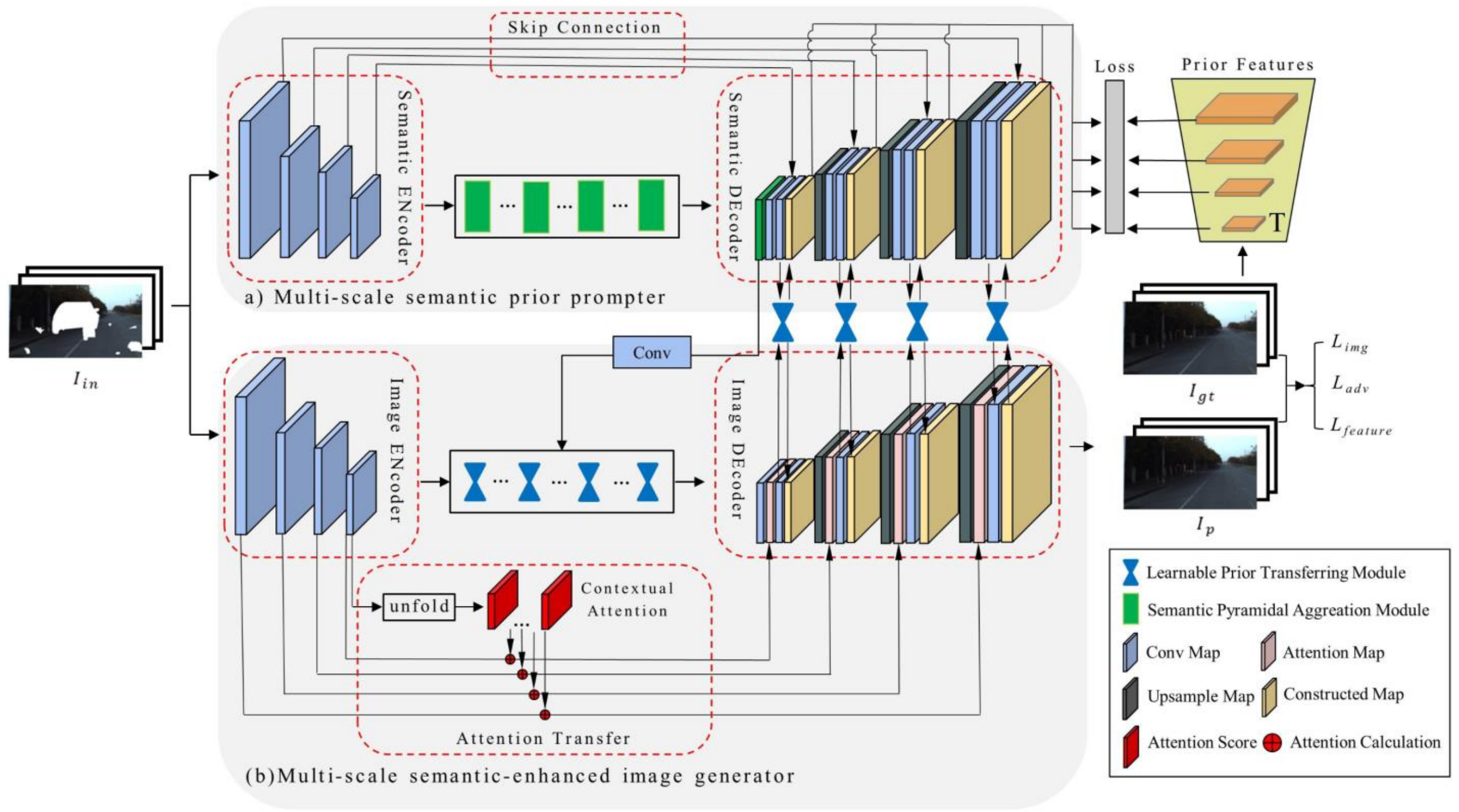

Fig. 2. Architecture of the proposed MFN: it is designed upon a dual Encoder-Decoder architecture, consisting of a multi-scale semantic prior prompter and a multi-scale semantic-enhanced image generator. The prompter learns multi-scale rich and discriminative representations under the supervision of pre-train model. The generator with LPT, a carefully designed module to adaptively fuse the semantic prior features and exchange information in decoder to restore reasonable structure and texture. The whole network is trained in an end-to-end manner.

### C. Visual prior-guided methods

With the continuous pursuit of image repair quality, the proportion of the visual prior-guided is gradually increasing, where a two-stage network[1] leveraged edge information as prior knowledge to enhance the consistency of an image's structure. Other approaches[30] also used edge-preserving images or semantic maps as extra supervision to achieve better performance in structure inpainting. Subsequently, the use of visual priors to guide the inpainting process has emerged as an active research area, with the goal of restoring structural information in images. Song et al.[20] employed semantic maps to guide the restoration of damaged images. However, the experiments only utilized small, rectangular-shaped masks, resulting color-bleeding effect in the boundary. Pinto et al.[19] introduced a multi-stage network that utilizes semantic maps and edges as prior information to enhance the structural consistency of image restoration. Similarly, Cao et al.[31] achieved successful restoration of structural details in indoor urban scenes (e.g., edges and lines) by mapping them into a specific tensor space. However, these methods with such successive procedures have limitations due to the fact that the restored image quality is highly dependent on the filled semantic map and edges, poorly filled prior will tremendously influence the restored images through error accumulation. Moreover, these methods require labor-intensive manual annotation for creating semantic maps, which can be challenging and time-consuming. In comparison, our method employs prior features learned at multiple scales from pre-trained models [32] to improve global context understanding. This approach provides more discriminative supervision during the inpainting process and reduces the need for heavy annotation in specific datasets. In addition, By adopting a single-stage and multi-scale feature fusion strategy in image restoration, our method effectively mitigates error accumulation and enhances the overall structural texture, particularly in high-resolution urban street-view images.

## III. METHODOLOGY

The deep generative model for street-view image inpainting is divided into parallel branches: multi-scale semantic prior prompter and multi-scale semantic-enhanced image generator, as illustrated in Figure 2. Both branches employ the Encoder-Decoder architecture for generation, along with the Soft-Mask Patch GAN[2] as a discriminator. In this section, we first present the multi-scale semantic prior prompter and supervision signals. Second, we provide a detailed description of multi-scale semantic-enhanced image generator. Next, we represent our experiment dataset and background-aware data processing. Finally, we elucidate the experimental training and testing strategy.

### *A. Multi scale semantic prior prompter*

According to Fig. 2(a), $I_{gt}$ indicates the ground truth, which is only involved in the training stage. Meanwhile, $I_{in}$ represents the damaged image, comprising the binary mask $M_{in}$ and $I_{gt}$. The workflow involves several key steps: initially, the input image $I_{in}$ and mask $M_{in}$ are fed into the semantic encoder, denoted as $E_s$, which extracts rich low-level features at different scales from the original input and generates a conditional input. Subsequently, we incorporate eight SPA modules, strategically enhancing the correlation between image content and semantic features. Finally, during the feature decoding stage, we exploit up-sampling layers to generate multi scale semantic priors with multi-scale skip connections. Simultaneously, we employ an image classification pre-trained model[32], denoted as $T$, to provide feature maps $S_l^{dist}$ as the supervision for learning the semantic priors. We aim to extract more discriminative representations of specific objects in complex street-view image, allowing the common knowledge to be transferred to the prior prompter. To concisely describe this process, we formulate (1-4) that capture the interplay of these components. All the size of feature map in our MFN could be represented as (5).

$$S_l^{enco} = E_s(I_{gt} * (1 - M_{in})) \tag{1}$$

$$S_L^{Deco} = \text{Conv}(\text{Multi_SPA_Modules}(S_L^{enco})) \tag{2}$$

$$S_l^{Deco} = \text{Conv}(\text{U}(S_{l+1}^{Deco}), S_l^{Enco}) \tag{3}$$

$$S_l^{dist} = T_s(I_{gt}) \tag{4}$$

$$\frac{h}{2^{l-1}} \times \frac{w}{2^{l-1}}, l \in \{1,2,3,4\}, L = 4 \tag{5}$$

Where $U$ donates the Up-sampling layer. The SPA and the loss function of prior promoter are described in detail next.

*1)Semantic Pyramidal Aggregation Module:* In the intricate milieu of complex street-view images, where objects vary in both scale and viewing angles, enhancing the network's semantic comprehension becomes imperative. This can be achieved by effectively capturing intricate relational patterns to facilitate contextual understanding. The structure of the SPA block is illustrated in Fig. 3. While dilated convolution[35] known for utilizing diverse sizes of dilated convolution kernels, enables the acquisition of global and local feature patterns. We have motivated to divide the standard convolution kernel into four sub-kernels on the channel and then overlay the feature maps obtained from smaller dilated convolution kernels layer-by-layer on each channel. This process enhances the reliability of global and local information, empowering the semantic prior prompter to capture a broad range of visual feature patterns and promote contextual content reasoning.

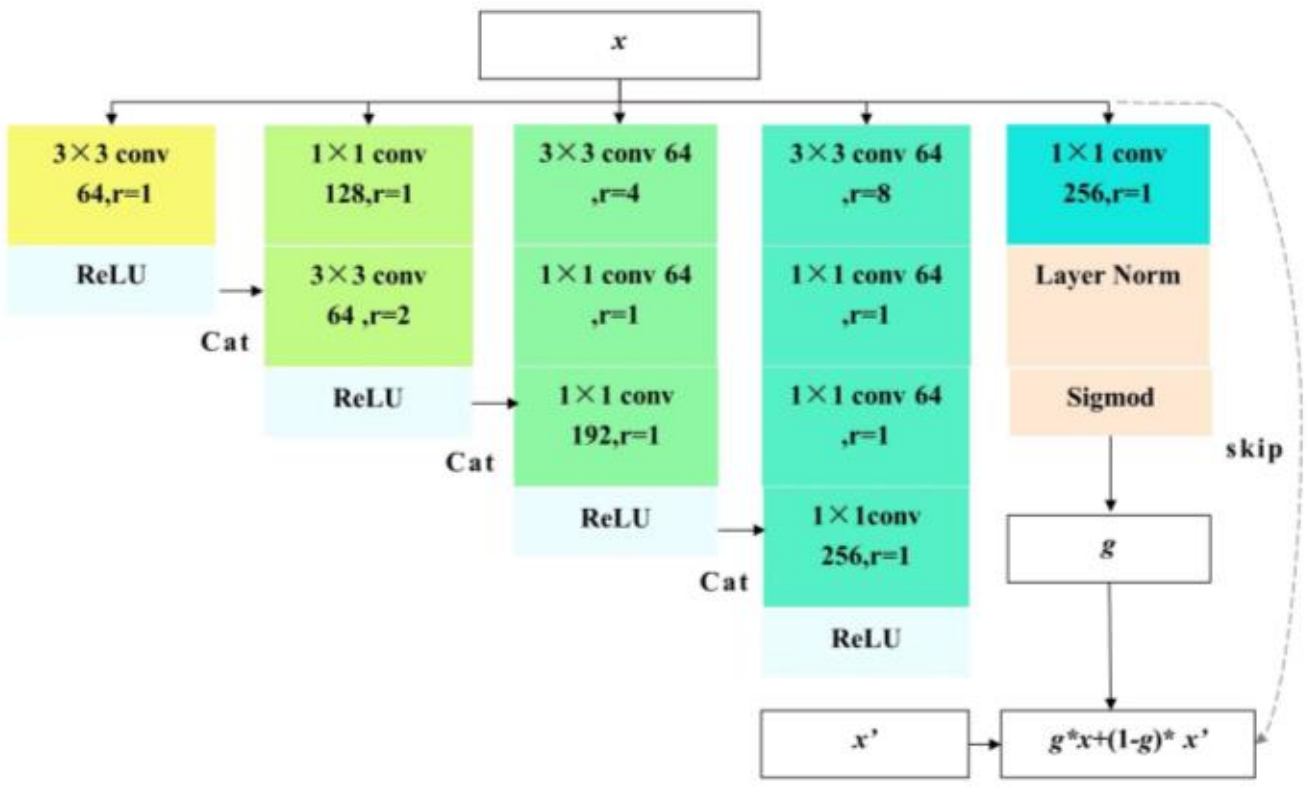

**Fig. 3.** An illustration of Our proposed *Semantic Pyramidal Aggregation* (SPA) Res Blocks. Cat(·) means the channel-wise concatenation.

In addition, the identical residual connections[33] can effectively alleviate gradient vanishing in deep networks, but they overlook the discrepancies in pixel values between the inside and outside regions of missing areas, leading to color disparity in the synthesized area. To address this issue, we employ gated residual connections [2], which further update the features inside the region while preserving the features outside the region. The gated residual connection employs a standard convolution operation on the input feature $x$ followed by a sigmoid function to calculate the gated value $g$. Subsequently, the SPA block performs a weighted sum using $g$ to aggregate the input feature $x$ and the learned residual feature $x'$.

$$x_{out} = x \times g + x' \times (1\text{-}g) \tag{6}$$

*2)Prior Loss:* Apart from the network architecture, another important focus is the efficient loss function. As we all know, the training process of prior promoter has certain guiding significance for image generator and can improve the stability of network. Therefore, we adopt the $L_1$ reconstruction loss[34] to supervise the semantic prior learner in learning specific prior features for filling missing regions:

$$L_{prior} = \sum_l \left\| (S_l^{dist} - \text{f}(S_l^{Deco}) \odot (1 + \alpha M_l)) \right\|_1 \tag{7}$$

The semantic prior loss incorporates the Hadamard product $\odot$ to calculate the error between the channel-mapped encoded features f() and the resized mask $M_l$, which has the same size as $S_l^{dist}$. Additionally, α represents an additional constraint for the masked region.

### *B. Multi scale semantic-enhanced image generator*

As shown in Fig. 2(b), we employ the same encoding strategy as mentioned in the semantic prior prompter to extract low-level visual representations from the damaged image $I_{in}$ and its binary mask $M_{in}$, but they do not share any parameters. Then we apply eight LPT modules to the last features $F_l^{enco}$ in image encoder to derive the $F_L^{Deco}$, aiming to adaptively incorporate the semantic priors $S_L^{Deco}$ into image encoding features. In addition, we adopt the attention transfer mechanism with all image encoding features to preserve the image details. During each level of image decoding stage, the attention map will be injected into image decoding feature $F_L^{Deco}$ to derive $F_l^{u}$. Then, the prior feature, $S_l^{Deco}$, and the up-sampled image feature $F_{l+1}^{u}$ of each decoding layer, are directed sent into LPT block for information exchange and adaptively fusion. In summary, we get the final results $I_p$ by (8-13).

$$F_l^{enco} = E_I(I_{gt} * (1 - M_{in})) \tag{8}$$

$$F_l^{Attn} = \text{AttentionTransfer}(F_l^{Enco}) \tag{9}$$

$$F_L^{Deco} = \text{Multi_LPT_Modules}(F_L^{enco}, \text{Conv}(S_L^{Deco})) \tag{10}$$

$$F_l^{u} = \text{Conv}(F_l^{Attn}, F_l^{Deco}) \tag{11}$$

$$F_l^{Deco} = \text{LPT}(\text{U}(F_{l+1}^{u}), S_l^{Deco}) \tag{12}$$

$$I_p = \text{Conv}(F_1^{Deco}) \tag{13}$$

where $F_l^{Attn}$ denotes the attention map. The LPT module, attention transfer mechanism and the final loss function of the image generator are described in detail next.

*1)Learnable Prior Transferring Module:* The reason for the LPT is that the image encoding feature and prior features, learned from the pre-trained model, exist in different domain spaces and feature levels. Concatenating or adding these features for straightforward fusion hampers the learning process. Therefore, there is a need to effectively combine these two branches of features into unified representations. The LPT block can adaptively incorporate learned prior features with the generated image features, as shown in Fig. 4. The LPT block serves two purposes: firstly, it facilitates the adaptive incorporation of multiscale semantic prior features and image features to generate coherent structures and clear textures. Secondly, it effectively exchanges semantic-image information in each layer of decoders to reduce error accumulation. To transfer the prior features into the image generator, the LPT block first normalizes the input image feature, $F_l^{'}$ using region normalization (RN). Subsequently, using SPADE, two sets of different affine transformation parameters, denoted as $[\beta, \gamma]$, and are learned from $S_l^{Deco}$ to conduct spatial pixel affine transformations on the image features $F_l^{'}$, as shown in (13) and (14).

$$[\beta, \gamma] = SPADE(S_l^{Deco}) \tag{13}$$

$$F_l^{'} = \beta \otimes \text{RN}(F_l^{u}) + \gamma \tag{14}$$

*2)Attention Transferring:* The rich low-level features play a crucial role in generating vivid textures and fine details during the inpainting process, such as texture and color. Hence, it is straightforward to incorporate a multi-level attention transfer mechanism that leverages images at the high-level features as a reference to capture low-level features and feed them into the decoder. Inspired by [35], we further adopt shared attention scores to reduce computational cost. Given a patch $f_i$ of fixed size learned from feature map $F_l^{enco}$, we first calculate its cosine similarity with another patch $f_j$. Then, we compute the attention scores $F_{i,j}^{soft}$ for $f_j$ *w.r.t* $f_i$ using a softmax function. Finally, we use the attention scores calculated by the last layer in the image encoder and then apply the attention scores back to the shallow layers to compute the attention maps. The whole process could be formulated by (15-17), where $N$ = 32*32=1024.

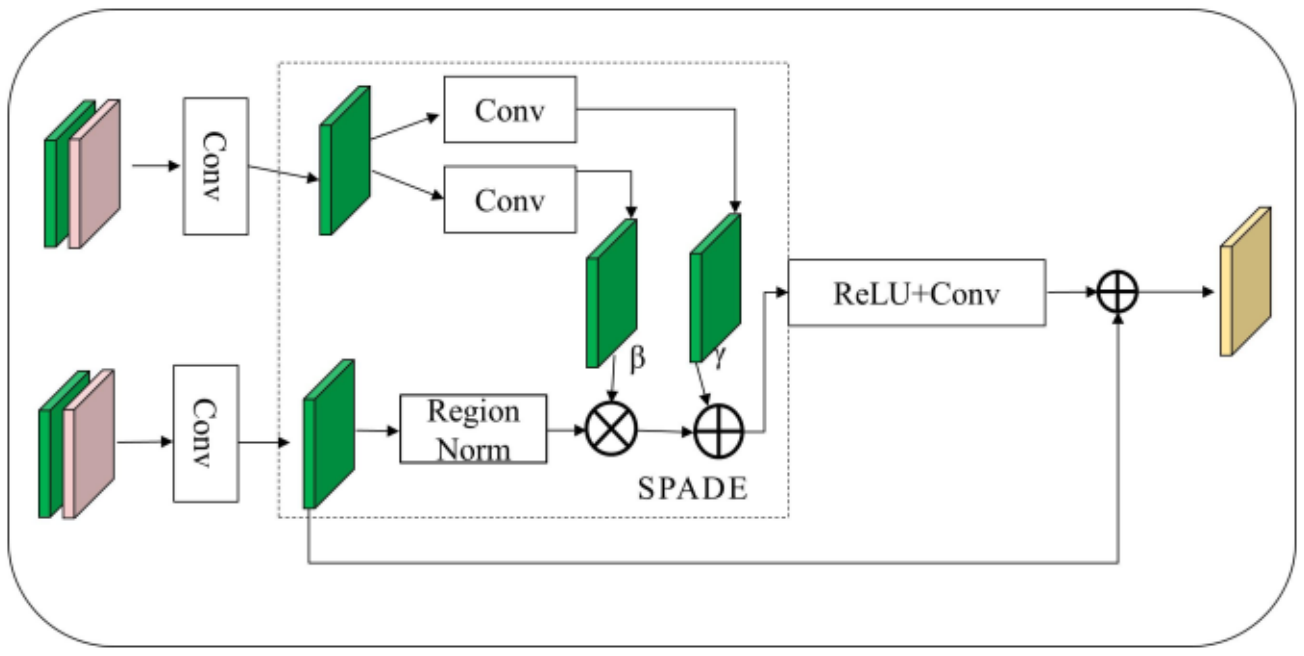

Fig. 4. Structures of our LPT Module and the Multi LPT Modules. The LPT consists one SPADE module to adaptively incorporate prior feature with image feature in image decoding-stage.

$$F_{ij} = \left\langle \frac{f_i}{\|f_i\|_2}, \frac{f_j}{\|f_j\|_2} \right\rangle \tag{15}$$

$$F_{i,j}^{soft} = \frac{\exp(F_{ij})}{\sum_{j=1}^{N} \exp(F_{ij})} \tag{16}$$

$$F_{Attn}^{l} = \sum_{j=1}^{N} F^{l} \bullet F_{i,j}^{soft} \tag{17}$$

*3)Final Loss Function:* The principle of choosing optimization objectives in image inpainting aims to generate accurate pixel-level reconstruction and high visual fidelity of the synthesized images. To this end, we employ several optimization objectives, including the $L_1$ loss, style loss[36] $L_{style}$, perceptual loss[37] $L_{perc}$, adversarial loss[2] $L_{adv}$, and the prior loss $L_{prior}$ mentioned earlier. Firstly, the $L_1$ loss is utilized to ensure precise reconstruction at the pixel level.

$$L_{rec} = ((I_p - I_{gt}) \odot (1 + \alpha M)) \tag{18}$$

The effectiveness of incorporating perceptual loss and style loss in image restoration tasks[35] has been widely demonstrated. The perceptual loss reduces the $L_1$ distance between the predicted and ground-truth value, enabling the network to preserve the structure and content of the original image.

$$L_{perc} = \sum_i \frac{\|\Phi_i(x) - \Phi_i(z)\|}{N_i} \tag{19}$$

where $\Phi_i(x)$ is the activation map from the *i*-th layers of a pre-trained network (e.g., VGG19[38]), $N_i$ is the number of elements in $\Phi_i(x)$ .The style loss measures the stylistic differences between the generated image and the original image. It also reduces the $L_1$ distance between the Gram matrix of the generated feature and the true values, allowing the generated image to better retain the texture and color of the original image.

TABLE I
DATASET DETAILS

| Name | Training | Image size for training | Testing | Image size for testing | Total |
|---|---|---|---|---|---|
| Apolloscapes | 20266 | 512×512 | 1200 | 1024×512 | 21466 |
| Cityscapes | 21600 | 512×512 | 1200 | 1024×512 | 22800 |

$$L_{style} = E\left[\left\|\Phi_i(x)^T\Phi_i(x) - \Phi_i(z)^T\Phi_i(z)\right\|_1\right] \quad (20)$$

Finally, in order to reduce the blurring in results, we employ the adversarial loss in SM-Patch GAN[2] to improve the visual fidelity of synthesized images.

$$L_{adv}^D = E_{z\sim P_z}\left[D(z) - \sigma(1-M))^2\right] + E_{x\sim P_{data}}\left[(D(x)-1)^2\right] \quad (21)$$

$$L_{adv}^G = E_{z\sim P_z}\left[(D(z)-1)^2 \odot M\right] \quad (22)$$

The full objective function of MFP can be written as:

$$L = L_{prior} + \lambda_{rec}L_{rec} + \lambda_{perc}L_{perc} + \lambda_{style}L_{style} + \lambda_{adv}L_{adv}^D \quad (23)$$

where the form of $L_{prior}$ is shown in (7). For our experiments, we empirically choose, $\alpha$ =1, $\lambda_{rec}$ =1, $\lambda_{perc}$ =0.5, $\lambda_{style}$ =250, $\lambda_{adv}$ =0.01 for training.

### *C. Dataset and background-aware data processing*

Recent studies have demonstrated that the augmentation of valid training data significantly enhances the performance of deep neural networks[39]. Therefore, we aim to obtain more valuable data by customizing the dataset. The image inpainting generative network requires learning texture from many un-occluded street-view images. Since there are many noise and undesirable pictures in the collected street view pictures, we propose a background-aware data processing to perform the sample image selection and encourage the generator to fill occluded regions with contextually plausible backgrounds. The pipeline consists of an image filter and mask producer, as shown in Fig. 6. Firstly, to create the image filter, we utilize the object detection model[40] trained on the large dataset[41] to filter out the images where the ratio of moving objects' areas is less than 5%. Such a filtering mechanism redirects the focus of generator towards background contents. Secondly, how to generate synthetic masks used during inpainting training also deserved to be considered. Because this will greatly affect whether the generated content by the generator is consistent with the background[42]. Specifically, we first pass the training images to object detection model[40] to generate highly accurate object bounding box. Next, we randomly sample some instance-level segmentation annotations from the internal database to create a hole, which serves as the initial mask along with object hole. Finally, we compute the overlapping ratio between the hole and each object bounding box from an image. If the overlapping ratio is larger than a threshold, we exclude the bounding box of each instance from the hole. Otherwise, we choose the bounding box of each instance as the output mask. And we set the threshold to 0.5 in this paper.

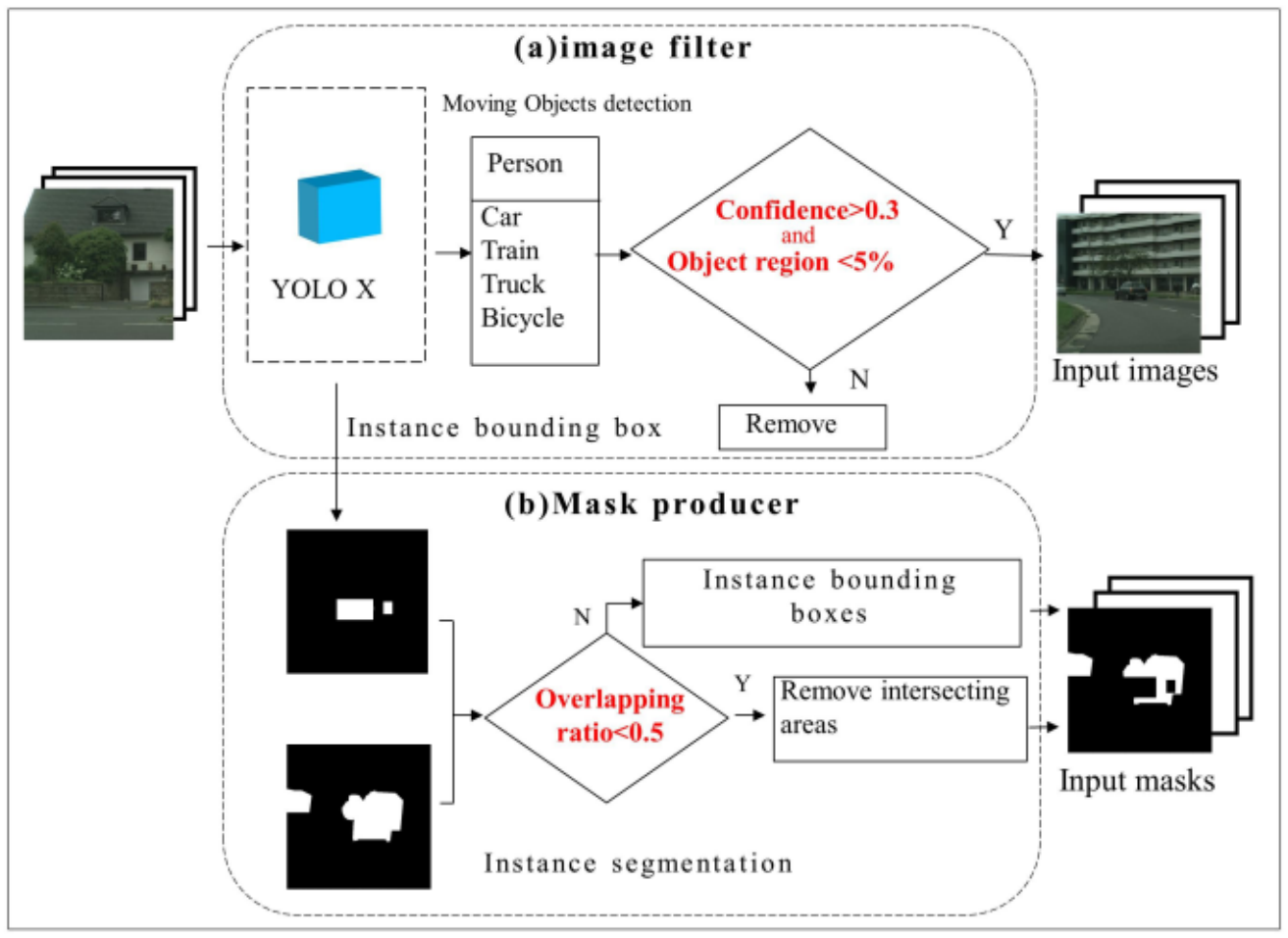

Fig. 6. An illustration of our proposed background-aware data processing pipeline. (a) The image filter with object detection model is purposed to filter out images with moving objects' area ratio is less than 5%. (b) The mask producer employs a filtering mechanism to generate object-aware masks that better simulate object removal cases.

We employ the Apolloscapes[43] dataset and the Cityscapes[44] dataset for training and testing in our experiments. The Cityscapes dataset is a comprehensive and high-quality collection of urban street scenes designed for advancing research in computer vision and deep learning. It encompasses a vast array of high-resolution images captured from various cities, showcasing diverse lighting scenarios, and traffic situations. The Apolloscapes dataset is a remarkable compilation of real-world data specifically curated to support research and development in the field of autonomous driving and map services. Both are crucial resources for training and evaluating algorithms related to scene understanding, autonomous driving, and urban environment analysis. We feed the image dataset into our data processing pipeline to generate image pairs for training and testing. The dataset details are listed in Table I. To conduct a fair evaluation, the images were randomly cropped into a size of 512×512 during the training phase, and a size of 1024×512 during the testing phase.

### *D. Training and evaluation*

The experiment is implemented with PyTorch under Ubuntu. When training network, we use the Adam with $\beta_1$ =0 and $\beta_2$ =0.99 as optimizer, the batch size is set to 8 and the maximum training iterations is 200K. The initial learning rate is set to 1×10-4 for all experiments, and we decay it 1×10-4 to 1×10-5 in the last 20K iterations. Besides, we also use the SM-Patch GAN[2] for adversarial training as in the previous work.

We conduct several comprehensive comparisons between our proposed method (MFN) and several state-of-the-art approaches that are specifically designed for high-resolution image inpainting. AOT [2] uses the transformer mechanism, which is successful in the field of deep learning in a single network. LaMa[3] use FFC to replace normal convolution, expanding receptive field in shallow network. MAT[18] design the mask-aware transformer block for large hole image inpainting. HINT[45] introduces the Mask-aware Pixel-shuffle Down-sampling Module (MPD) and Spatially-activated Channel Attention Layer to preserve visible information from corrupted images, especially in scenarios with extensive missing regions. To fairly assess the performance of these networks, we retrain the models using the provided code and evaluate them using the same image-mask pairs to verify the accuracy of the networks. In terms of evaluation indicators, we employ peak signal-to-noise ratio[18, 29] (PSNR), structural similarity[46] (SSIM) , mean absolute error[27, 47] (MAE), root mean square error[47] (RMSE), and Learned Perceptual Image Patch Similarity(LPIPS) to assess the quality of the synthesized image.

To the best of our knowledge, there is currently no dataset available that offers a specific and precise collection of paired static and dynamic street-view images. As a result, assessing the quality of synthesized images after object removal using objective metrics becomes a challenge. Therefore, a subjective visual comparison survey is further conducted to measure the performance of our proposed method. We invited 30 experts in image processing to participate in the survey. The expert, unaware of the specific inpainting method under evaluation, are tasked with selecting the most plausible image from our proposed approach and baseline methods. Each expert answers 40 randomly selected questions from either the Apolloscapes or Cityscapes dataset. We collect 1300 valid votes to ensure reliable statistical analysis, diligently recording participants' responses and summarizing the resulting statistical data.

## IV. RESULT ANALYSIS

### *A. Object Removal Qualitative Comparison*

The Object removal qualitative comparisons on the Apolloscapes dataset and Cityscapes dataset between the baseline methods and our method in Fig. 6 and Fig. 7. Compared to the baseline methods, the significant improvement of our method is the more reasonable and fine-grained texture. The multi-scale semantic prior features provide rational guidance for generating intricate scene targets and edges, thereby enabling superior detection of the structure and color of the missing areas. The input images from the Apolloscape dataset are shown in Fig. 6. These images showcase urban street views with varying lighting conditions, locations, and levels of occlusion caused by moving objects. The synthesized results of the AOT are displayed in Fig. 6(b). All results are failed to capture the reasonable structure and significantly affected by the shadow, resulting in distorted and visually hazy content in the hole. The restored results of LaMa and MAE are presented in Fig. 6(c) and Fig. 6(d). Although both can accurately reconstruct the overall texture and structure, they fail to capture fine details in the yellow regions, leading to partial artifacts. HINT has shown state of art performance on the Place and Cereb datasets, but the results obtained Apolloscape dataset, which contains more complex urban scene, still show serious distortions in structure, as shown in Fig. 6(e). In contrast, our method generates more reasonable global structures and more distinct image textures, such as a more natural road boundary and the shadow on the road. These can strongly validate that our method excels in synthesizing natural structures and textures that align with human imagination.

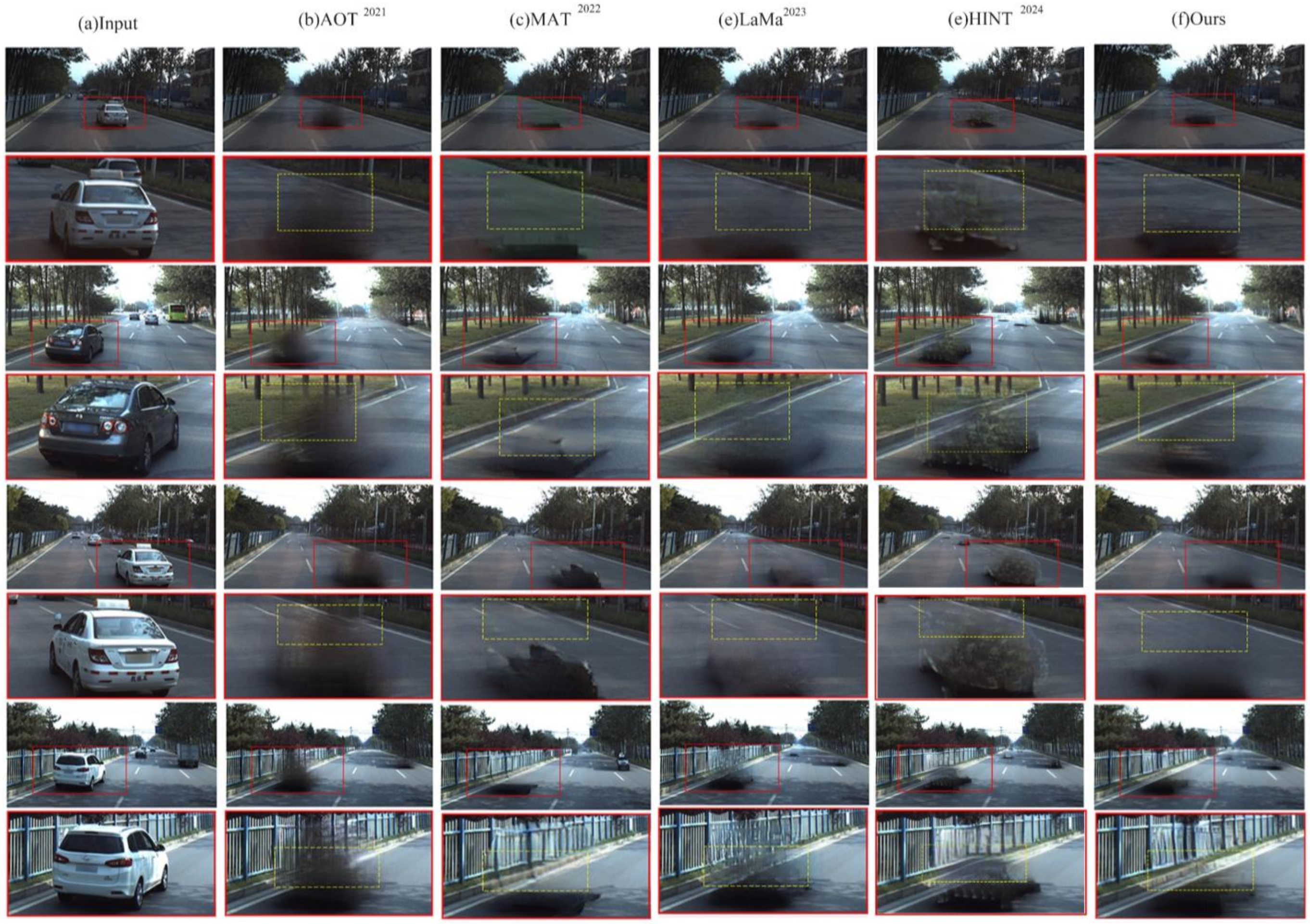

Fig. 6. Comparisons with existing methods on Apolloscapes dataset. The area in red box is enlarged for better visualization.

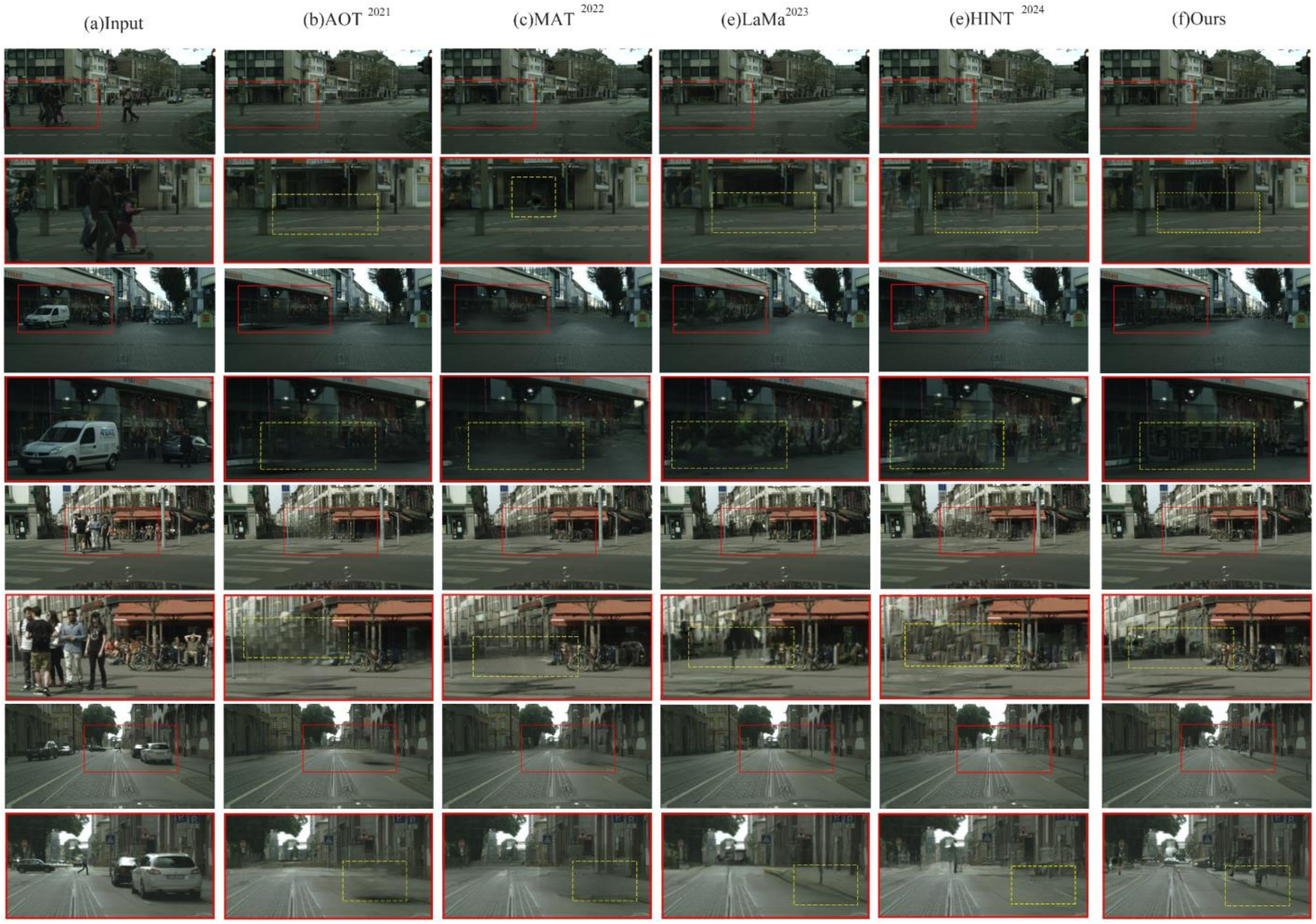

Fig. 7. Comparisons with existing methods on Cityscapes dataset. The area in red box is enlarged for better visualization.

The Apolloscape dataset primarily consists of urban highways captured exclusively in Beijing with minimal variation in data distribution. In contrast, The Cityscapes dataset presents a myriad of complex and diverse scenes, characterized by significant variations in lighting conditions and a multitude of object label categories. Therefore, conducting the moving objects removal task on it poses a significantly greater challenge. In the first and third row of Fig 7, the overall images are dark and portray a densely populated urban street intersection. The restored results by AOT and MAT all present a small amount of mask traces and the textures seem blurry, while the result of LaMa fails to complete the structure of the door in the yellow regions. In the fifth and seventh rows of Fig. 7, the light condition of the images is brighter and occluded regions are bigger than the upper images. The results of AOT, LaMa and HINT failed to generate building and road structures, filling in the holes with checkerboard artifacts. The MAT generates distorted building and road structures and blurry details. While our network accurately restores identification lines and road shoulders and well preserves fine-grained textures.

### B. Inpainting Quantitative Comparison

To conduct a more detailed comparison, the mask set are pre-grouped into six intervals according to the size of the masks (0%-10%, 10%-20%, . . ., 50%-60%). As illustrated in Table III and Table IV, the results highlight the superiority of our proposed network in comparison to existing methods. Especially, even when confronted with challenging urban street view images containing a substantial proportion of missing areas (i.e., 40-60% mask), our network exhibits excellent reconstruction capability due to its global understanding of the image. In Cityscapes datasets, we can clearly observe the significant improvements achieved by our proposed network across all mask rates on all metrics. Our algorithm improves the PSNR and SSIM by 3.08% and 0.90%, respectively, and meanwhile reduces the MAE, RMSE, and LPIPS by 16.72%, 7.83%, and 19.24%, respectively, compared to the LaMa at large hole-to-image 60% mask. Compared to HINT, we also improved MAE and LPIPS by nearly 9.5% and 41.07% on average. In Apolloscape dataset, despite the relatively fixed city highway structure, the proposed method still improves the MAE, RMSE and LPIPS by 16.71%, 2.46% and 3.0%, respectively, compared to LaMa with the 60% mask. Overall, MFN increase the PSNR and SSIM by about 0.40-1.90% and 0.28%-1.22%, and decreases the MAE, RMSE and LPIPS by 8.07%-20.10%, 1.35%-14.41% and 2.37%-16.71% on Apolloscape dataset. While on Cityscapes dataset, MFN achieves better performance, which increase the PSNR and SSIM by about 0.24-4.33% and 0.46%-1.24%, and decreases the MAE, RMSE and LPIPS by 9.55%-19.71%, 0.19%-16.89% and 16.69%-41.79%.

TABLE III
QUANTITATIVE COMPARISON BETWEEN THE BASELINE METHODS AND OUR METHOD ON APOLLOSCAPE DATASET ( ↓ LOWER IS BETTER. ↑ HIGHER IS BETTER. BEST RESULTS ARE HIGHLIGHTED).

| Mask Ratio | Model | PSNR | SSIM | MAE | RMSE | LPIPS |
|---|---|---|---|---|---|---|
| 10% | AOT2021 | 35.35 | 0.9653 | 0.0069 | 5.70 | 0.0309 |
| | MAT2022 | 35.60 | 0.9713 | 0.0063 | 4.88 | 0.0286 |
| | LaMa2023 | 35.70 | 0.9723 | 0.0048 | 4.85 | 0.0244 |
| | HINT2024 | 35.49 | 0.9725 | 0.0049 | 5.00 | 0.0250 |
| | OURS | **35.82** | **0.9728** | **0.0055** | **4.84** | **0.0243** |
| 20% | AOT2021 | 30.43 | 0.9117 | 0.0132 | 9.62 | 0.0690 |
| | MAT2022 | 30.71 | 0.9275 | 0.0126 | 8.94 | 0.0623 |
| | LaMa2023 | 30.81 | 0.9280 | 0.0112 | 8.22 | 0.0556 |
| | HINT2024 | 30.64 | 0.9283 | 0.0110 | 8.19 | 0.0949 |
| | OURS | **30.84** | **0.9297** | **0.0108** | **8.15** | **0.0554** |
| 30% | AOT2021 | 27.21 | 0.8791 | 0.0185 | 13.51 | 0.1025 |
| | MAT2022 | 27.62 | 0.8835 | 0.0174 | 11.89 | 0.0941 |
| | LaMa2023 | 27.68 | 0.8859 | 0.0163 | 11.54 | 0.0864 |
| | HINT2024 | 27.65 | 0.8860 | 0.0164 | 11.57 | 0.0877 |
| | OURS | **27.76** | **0.8870** | **0.0147** | **11.32** | **0.0850** |
| 40% | AOT2021 | 26.73 | 0.8477 | 0.0226 | 13.81 | 0.1267 |
| | MAT2022 | 27.11 | 0.8542 | 0.0214 | 12.28 | 0.1145 |
| | LaMa2023 | 27.25 | 0.8515 | 0.0218 | 11.97 | 0.1061 |
| | HINT2024 | 27.28 | 0.8546 | 0.0220 | 11.45 | 0.1161 |
| | OURS | **27.41** | **0.8553** | **0.0182** | **11.78** | **0.1040** |
| 50% | AOT2021 | 25.39 | 0.8131 | 0.0297 | 16.61 | 0.1491 |
| | MAT2022 | 25.74 | 0.8154 | 0.0274 | 14.90 | 0.1377 |
| | LaMa2023 | 25.77 | 0.8178 | 0.0276 | 14.42 | 0.1307 |
| | HINT2024 | 25.53 | 0.8128 | 0.0243 | 14.48 | 0.1389 |
| | OURS | **25.80** | **0.8227** | **0.0237** | **14.28** | **0.1257** |
| 60% | AOT2021 | 24.80 | 0.7784 | 0.0370 | 16.69 | 0.1673 |
| | MAT2022 | 25.19 | 0.7828 | 0.0342 | 15.06 | 0.1545 |
| | LaMa2023 | 25.24 | 0.7878 | 0.0359 | 14.98 | 0.1475 |
| | HINT2024 | 25.34 | 0.7831 | 0.0326 | 15.19 | 0.1453 |
| | OURS | **25.51** | **0.7910** | **0.0293** | **14.62** | **0.1432** |
| Average | AOT2021 | 28.32 | 0.8659 | 0.0213 | 12.66 | 0.1076 |
| | MAT2022 | 28.66 | 0.8725 | 0.0199 | 11.33 | 0.0986 |
| | LaMa2023 | 28.74 | 0.8739 | 0.0196 | 11.00 | 0.0918 |
| | HINT2024 | 28.65 | 0.8729 | 0.0185 | 10.98 | 0.1013 |
| | OURS | **28.86** | **0.8764** | **0.0170** | **10.83** | **0.0896** |

### *C. User Study*

The results of visual comparison survey suggest that the proposed MFN significantly improves the reliability and validity of street view, as shown in Table V. Results show that our method receives most votes on realistic object removal requests, with all support rates exceeding 70%. And it's worth noting that in comparison with the results of the MAT algorithm, the MFN algorithm obtained a high support rate of 81.86%. The subjective assessment by domain experts, who were unbiased regarding the specific method employed, reinforces the conclusion that our method shows more advantages and efficiency.

TABLE IV
QUANTITATIVE COMPARISON BETWEEN THE BASELINE METHODS AND OUR METHOD ON CITYSCAPES DATASET ( ↓ LOWER IS BETTER. ↑ HIGHER IS BETTER. BEST RESULTS ARE HIGHLIGHTED).

| Mask Ratio | Model | PSNR | SSIM | MAE | RMSE | LPIPS |
|---|---|---|---|---|---|---|
| 10% | AOT2021 | 32.40 | 0.9593 | 0.0063 | 7.89 | 0.0369 |
| | MAT2022 | 32.50 | 0.9641 | 0.0069 | 7.17 | 0.0368 |
| | LaMa2023 | 32.57 | 0.9653 | 0.0058 | 7.06 | 0.0336 |
| | HINT2024 | 32.65 | 0.9632 | **0.0038** | **5.42** | 0.0516 |
| | OURS | **32.77** | **0.9679** | 0.0055 | 6.75 | **0.0302** |
| 20% | AOT2021 | 27.04 | 0.9149 | 0.0132 | 13.02 | 0.0766 |
| | MAT2022 | 27.74 | 0.9207 | 0.0126 | 11.69 | 0.0730 |
| | LaMa2023 | 28.04 | 0.9214 | 0.0112 | 11.25 | 0.0740 |
| | HINT2024 | 28.34 | 0.9217 | **0.0099** | **9.99** | 0.1082 |
| | OURS | **28.46** | **0.9264** | 0.0108 | 10.84 | **0.0608** |
| 30% | AOT2021 | 26.03 | 0.8785 | 0.0185 | 15.93 | 0.1103 |
| | MAT2022 | 26.42 | 0.8820 | 0.0174 | 14.34 | 0.1094 |
| | LaMa2023 | 26.79 | 0.8868 | 0.0163 | 13.99 | 0.1069 |
| | HINT2024 | 26.67 | 0.8841 | 0.0163 | 13.62 | 0.1570 |
| | OURS | **27.13** | **0.8895** | **0.0147** | **12.93** | **0.0873** |
| 40% | AOT2021 | 24.44 | 0.8488 | 0.0226 | 17.86 | 0.1380 |
| | MAT2022 | 24.83 | 0.8488 | 0.0218 | 17.30 | 0.1356 |
| | LaMa2023 | 25.00 | 0.8513 | 0.0214 | 16.21 | 0.1276 |
| | HINT2024 | 25.62 | 0.8538 | 0.0193 | **14.29** | 0.1903 |
| | OURS | **25.68** | **0.8552** | **0.0182** | 15.00 | **0.1057** |
| 50% | AOT2021 | 22.34 | 0.8024 | 0.0297 | 21.47 | 0.1684 |
| | MAT2022 | 22.93 | 0.8085 | 0.0276 | 19.95 | 0.1659 |
| | LaMa2023 | 23.13 | 0.8096 | 0.0274 | 19.32 | 0.1560 |
| | HINT2024 | 23.43 | 0.8091 | 0.0257 | 18.56 | 0.2234 |
| | OURS | **23.85** | **0.8144** | **0.0237** | **17.86** | **0.1299** |
| 60% | AOT2021 | 21.95 | 0.7658 | 0.037 | 24.37 | 0.2002 |
| | MAT2022 | 22.11 | 0.7769 | 0.0359 | 23.47 | 0.1983 |
| | LaMa2023 | 22.29 | 0.7734 | 0.0342 | 21.75 | 0.1822 |
| | HINT2024 | 22.78 | 0.7776 | 0.038 | 21.83 | 0.2431 |
| | OURS | **23.00** | **0.7805** | **0.0293** | **20.17** | **0.1528** |
| Average | AOT2021 | 25.70 | 0.8616 | 0.0212 | 16.76 | 0.1217 |
| | MAT2022 | 26.09 | 0.8668 | 0.0204 | 15.65 | 0.1198 |
| | LaMa2023 | 26.30 | 0.8680 | 0.0194 | 14.93 | 0.1134 |
| | HINT2024 | 26.75 | 0.8683 | 0.0188 | 13.95 | 0.1623 |
| | OURS | **26.82** | **0.8723** | **0.0170** | **13.93** | **0.0945** |

## D. Ablation Studies

In this section, we aim to investigate the contributions of different components within our framework to the overall performance and assess the impact of using different supervised models on the generated results. All models are trained as well as tested using the same settings for fair comparisons. The above research is conducted on the Cityscapes dataset.

*1)The influence of each module of the network:* We conduct quantitative comparisons, as shown in Table VI, to examine the influence of each module of the network. Firstly, we show the contributions of the semantic supervision by removing it from the network and directly concatenating the learned semantic priors and the image features in decoding stage, showing the substantial performance degradation. From the comparative

TABLE V

USER STUDY FOR PAIR-WISE COMPARISONS WITH AOT, MAT, AND LAMA ON APOLLOSCAPES AND CITYSCAPES. “>” MEANS OUR RESULT IS BETTER THAN THE OTHER. “<” MEANS OUR RESULT IS WORSE THAN THE OTHER. “=” MEANS OUR RESULT IS EQUAL TO THE OTHER.

| Dataset | Apolloscapes | | | Cityscapes | | |
|---|---|---|---|---|---|---|
| Votes | > | < | = | > | < | = |
| Ours vs AOT | **80.39 %** | 8.82 % | 10.78 % | **85.24 %** | 6.19 % | 8.57 % |
| Ours vs MAT | **81.86 %** | 11.50 % | 6.64 % | **82.95 %** | 11.06 % | 5.99 % |
| Ours vs LaMa | **70.91 %** | 12.27 % | 16.82 % | **75.34 %** | 12.56 % | 12.11 % |

TABLE VI

TABLE 4. COMPARISONS ON THE CITYSCAPES DATASET OF (A) FULL MODEL (B) WITHOUT SEMANTIC SUPERVISION (C) WITHOUT LPT MODULE IN IMAGE GENERATOR. (D) WITHOUT MULTI-SCALE STRUCTURE IN PRIOR PROMPTER. (E) WITHOUT ATTENTION TRANSFER (F)WITHOUT SPA MODULE IN PRIOR PROMPTER.

| Type | Model | PSNR | SSIM | MAE | RMSE | LPIPS |
|---|---|---|---|---|---|---|
| **A** | **Full Model** | **26.81** | **0.8723** | **0.0170** | **13.92** | **0.0945** |
| B | - W/O Semantic Supervision | 25.98 ↓ 0.83 | 0.8611 ↓ 0.0112 | 0.0250 ↑ 0.0080 | 15.56 ↑ 1.64 | 0.1233 ↑ 0.0288 |
| C | - W/O LPT Block | 26.06 ↓ 0.75 | 0.8636 ↓ 0.0087 | 0.0239 ↑ 0.0069 | 15.02 ↑ 1.10 | 0.1217 ↑ 0.0272 |
| D | - W/O Multi-scale Structure | 26.21 ↓ 0.60 | 0.8690 ↓ 0.0033 | 0.0224 ↑ 0.0054 | 14.62 ↑ 0.70 | 0.1147 ↑ 0.0202 |
| E | - W/O Attention Transfer | 26.34 ↓ 0.47 | 0.8702 ↓ 0.0021 | 0.0193 ↑ 0.0031 | 14.38 ↑ 0.46 | 0.1077 ↑ 0.0132 |
| F | - W/O SPA Block | 26.66 ↓ 0.15 | 0.8706 ↓ 0.0017 | 0.0182 ↑ 0.0042 | 14.12 ↑ 0.20 | 0.1023 ↑ 0.0078 |

results shown in rows A and B, we could conclude that the reduced model could not effectively capture helpful semantic information for global context understanding. Additionally, to evaluate the effects of the LPT module, we directly concatenate the prior features and image features, and then the results in row C demonstrate that the LPT module could adaptively fuse multi-scale semantic priors and low-level features. Besides, we show the effectiveness of leveraging multi-scale features in the prompter by only integrating with the smallest spatial size (64 × 64) prior into the image generator. As shown in row D, we observe a significant decline in performance which demonstrates the importance of the incorporation of multi-scale features in image inpainting. Furthermore, we replace the attention transfer by skip connection and obtain the result in row E. The obvious performance decline proves that attention map in image inpainting facilitates the preservation of coherence and fine details in the inpainted regions. Finally, we removed the SPA module from the semantic prior prompter, leading to the results shown in row F. The performance decline further emphasizes the significance of enriched visual feature patterns in facilitating contextual content reasoning.

*2)The different supervision:* Considering that other types of priors, such as semantic maps and edge maps, have been employed in existing street-view image inpainting methods, and we explore alternative supervisions for learning prior features and determine whether they could enhance global understanding. We analyze the structural information generated by the EC[1] (Edge Completion) model, as illustrated in Figure 8(a). We observe that the synthesized edges failed to restore reasonable structures from the local to the global region, resulting in significant artifacts in synthesized urban street-view images. Subsequently, we visualize the multi-level learned priors under different semantic supervision, including object classification model TResNet[32] trained on the Open Image dataset[48], a target detection model YoloV7[49] trained on the COCO[41] dataset and a semantic segmentation model HRNet[50] trained on the ADE20K[51] dataset. By clustering the prior features learned at different scales using the ISODATA algorithm, we obtained the results shown in Figure 8(b). The result indicates that the semantic prior prompter performed better under the supervision of the object classification model. Because the object classification model provides valuable information on global semantic understanding. Semantic information involves the objects in the image, their attributes, spatial relationships, and the overall structure of the scene. While the object detection model focuses more on foreground targets, resulting in features that emphasize their location. And the semantic segmentation model primarily emphasizes the boundary of the target objects rather than the semantic understanding of complex scenes.

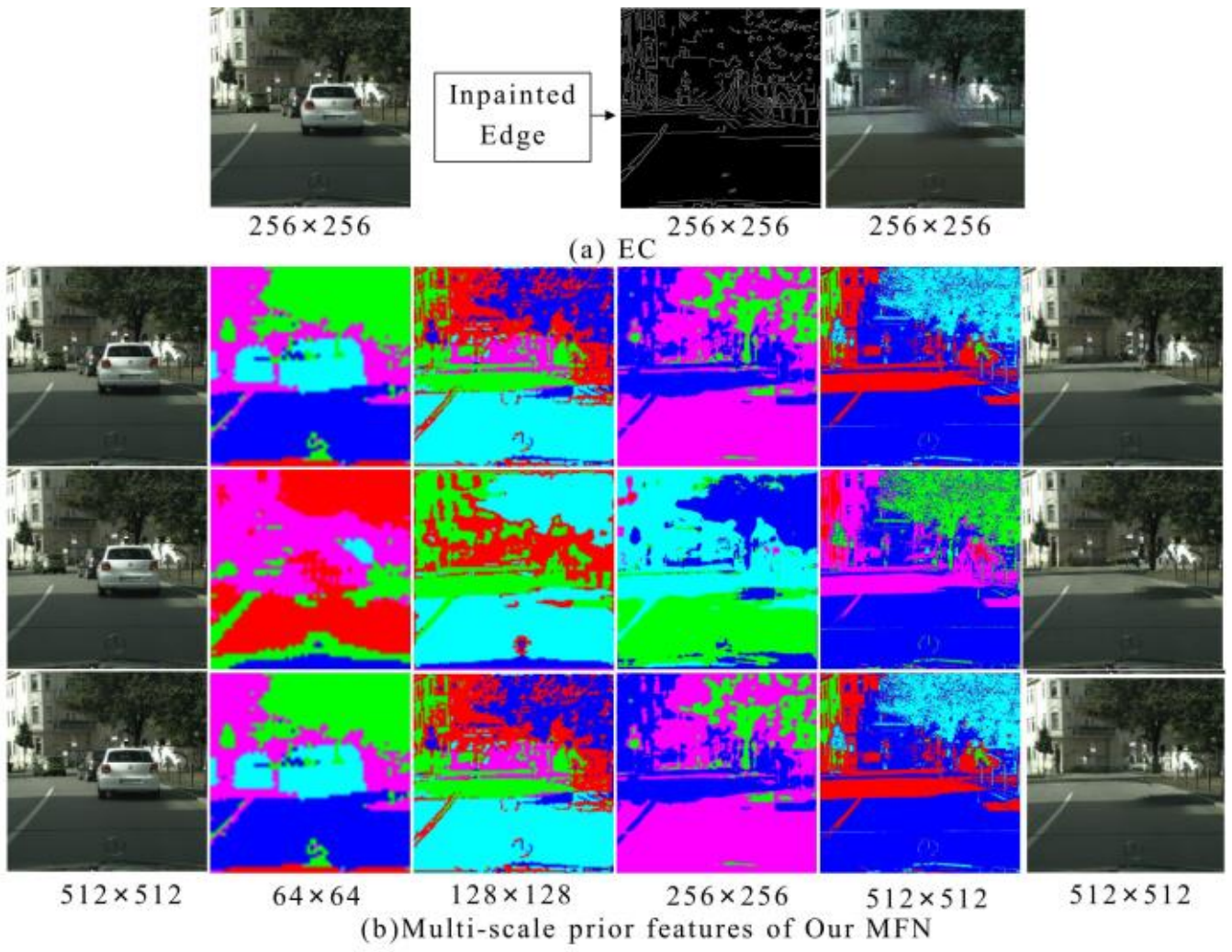

**Fig. 8.** Comparisons of the learned prior features. In (a), the figure shows the synthesized edge map and image in EC. In (b), the figure shows the multi-scale prior features extracted from our MPF. The results of each row are supervised by different pretext models, which are listed as follows: TResNet for a classification task trained on the Open Image dataset, YoloV7 for object detection trained on the COCO dataset, and HRNet for semantic segmentation trained on the ADE20K dataset.

According to the above experimental results, our algorithm achieves the highest evaluation index and ensures the stability of network repair results in a large number of experiments, in terms of the quantitative, qualitative evaluation and visual comparison survey. With the guidance of semantic priors and attention map injected, we improve performance of our MPF model in comprehending global content, reasoning missing structural information, and generating visually realistic synthesized results.

## V. CONCLUSION

In this paper, we propose a specifically designed framework named MFN which could generate street-view images without any moving objects. To achieve global context understanding of an image, a semantic prior prompter is proposed with SPA blocks to learn multi-scale and discriminative representations from the pretext model. In addition, a multi-scale semantic-enhanced image generator with LPT blocks is designed not only for adaptively fusing the semantic prior features with low-level image features, but also for effectively exchanging semantic-image information in each layer of decoders. Finally, a background-aware data processing pipeline is developed to prevent the generator from synthesizing new objects inside the hole and reduce color-bleeding effects across the boundaries, meeting the object removal in real-world use cases. We conduct experiments on different resolutions and types of data to verify the effectiveness of the proposed model. In the experimental results of Apolloscapes dataset and the Cityscapes dataset, our algorithm shows improvements of approximately 1.9%, 0.4% in PSNR, 13.09%,12.12% in MAE and 2.37%, 16.89% in LPIPS, respectively compared to the baseline approaches. Simultaneously, a succession of supplementary ablation experiments has been conducted to affirm the efficacy of our proposed Multi-Feature Network (MFN). In conclusion, our method undeniably presents a compelling resolution, not solely for safeguarding privacy, but also for cultivating more dependable datasets tailored to urban applications involving street view imagery. Future work includes integrating our algorithm into advanced architectures and leveraging additional datasets to enhance its robustness and contribute to the advancement of the research area. These efforts hold the potential to be of great value to the broader scientific community.

## ACKNOWLEDGMENT

The authors would like to thank the authors of the AOT, MAT, and LaMa for their efforts in sharing the code. They would also like to thank Huang X et al. and Cordts M et al. for providing the Apollopscapes dataset and Cityscapes dataset used in this study.